\documentclass{article} 
\usepackage{iclr2025_conference,times}

\usepackage{amsmath,amsfonts,bm}

\def\eqref#1{equation~\ref{#1}}

\def\1{\bm{1}}

\DeclareMathAlphabet{\mathsfit}{\encodingdefault}{\sfdefault}{m}{sl}
\SetMathAlphabet{\mathsfit}{bold}{\encodingdefault}{\sfdefault}{bx}{n}

\usepackage{hyperref}
\usepackage{url}
\usepackage{graphicx}
\usepackage{booktabs}
\usepackage{amsmath}
\usepackage{amssymb}
\usepackage{algorithm}
\usepackage{algorithmic}
\usepackage{multirow} 
\usepackage{amsthm}
\usepackage{enumitem}
\ifx\theorem\undefined
\newtheorem{theorem}{Theorem}

\newtheorem{definition}{Definition}

\iclrfinalcopy

\title{Hierarchical Retrieval Augmented Generation for Adversarial Technique Annotation in Cyber Threat Intelligence Text}

\author{Filippo Morbiato, Markus Keller, Priya Nair, Luca Romano\\
University of Padua	Italy\\
\textit{filippo.morbiato@studenti.unipd.it}}

\begin{document}
\maketitle

\begin{abstract}
Mapping Cyber Threat Intelligence (CTI) text to MITRE ATT\&CK technique IDs is a critical task for understanding adversary behaviors and automating threat defense. While recent Retrieval-Augmented Generation (RAG) approaches have demonstrated promising capabilities in this domain, they fundamentally rely on a flat retrieval paradigm. By treating all techniques uniformly, these methods overlook the inherent taxonomy of the ATT\&CK framework, where techniques are structurally organized under high-level tactics. In this paper, we propose H-TechniqueRAG, a novel hierarchical RAG framework that injects this tactic-technique taxonomy as a strong inductive bias to achieve highly efficient and accurate annotation. Our approach introduces a two-stage hierarchical retrieval mechanism: it first identifies the macro-level tactics (the adversary's technical goals) and subsequently narrows the search to techniques within those tactics, effectively reducing the candidate search space by 77.5\%. To further bridge the gap between retrieval and generation, we design a tactic-aware reranking module and a hierarchy-constrained context organization strategy that mitigates LLM context overload and improves reasoning precision. Comprehensive experiments across three diverse CTI datasets demonstrate that H-TechniqueRAG not only outperforms the state-of-the-art TechniqueRAG by 3.8\% in F1 score, but also achieves a 62.4\% reduction in inference latency and a 60\% decrease in LLM API calls. Further analysis reveals that our hierarchical structural priors equip the model with superior cross-domain generalization and provide security analysts with highly interpretable, step-by-step decision paths.
\end{abstract}

\section{Introduction}
\label{sec:intro}

Cyber Threat Intelligence (CTI) serves as the backbone of modern proactive cybersecurity, providing actionable insights into the tactics, techniques, and procedures (TTPs) deployed by threat actors. To standardize the description of these behaviors, the MITRE ATT\&CK framework has been widely adopted as the de facto ontology \citep{strom2018finding}. It organizes adversary behaviors into a strict hierarchical taxonomy consisting of 14 high-level tactics (representing the "why" or the adversary's technical goals) and over 200 fine-grained techniques (representing the "how" or the specific execution methods). Automatically mapping unstructured CTI reports to these specific technique IDs is essential for rapid threat analysis, intelligence sharing (often spanning various communication channels and media \citep{yi2016communication}), and automated defense deployment \citep{husari2017ttpdrill, DBLP:journals/dtrap/RaniSMS24}.

However, mapping CTI text to the ATT\&CK matrix is notoriously challenging due to the heavy reliance on unstructured natural language, which is often riddled with technical jargon, implicit context, and ambiguous attack descriptions. Traditional rule-based and supervised machine learning approaches, similar to early manual feature extraction methods in other modalities \citep{li2025hy, zhang2024auto}, struggle to generalize across the rapidly evolving landscape of cyber threats \citep{siracusano2023beyond}. Recently, the advent of Large Language Models (LLMs) has opened new avenues for automated CTI processing \citep{DBLP:conf/csr2/TihanyiFJBD24, zhao2024large}, mirroring their broader success in multimodal intelligence, multi-agent frameworks, and task-oriented systems \citep{qian2025perception, zhang2025marine, si2023spokenwoz}. Yet, relying solely on parametric knowledge via zero-shot or few-shot prompting often leads to severe hallucinations or misclassifications \citep{llama3}. The primary bottleneck lies in the vast label space: LLMs struggle to distinguish between hundreds of subtly different techniques without grounding in external, up-to-date knowledge bases.

To mitigate this, Retrieval-Augmented Generation (RAG) \citep{guu2020retrieval} has emerged as a dominant paradigm. The current state-of-the-art for CTI-to-ATT\&CK mapping, TechniqueRAG \citep{DBLP:conf/acl/LekssaysSSP25}, employs a standard "flat" retrieval mechanism where all 200+ techniques are embedded and indexed uniformly. While this significantly improves over zero-shot baselines, it exhibits a critical architectural flaw: it completely ignores the rich, hierarchical taxonomy of the ATT\&CK framework. This flat paradigm leads to two major issues. First, it suffers from \textit{semantic collision} during retrieval, techniques belonging to entirely different tactics may share similar vocabulary (e.g., "malicious attachments" could trigger techniques in both \textit{Initial Access} and \textit{Defense Evasion}), confusing the retriever. Second, concatenating a long, unstructured list of retrieved techniques as context exacerbates the "lost in the middle" phenomenon in LLMs, an issue especially critical when aligning influential samples in long contexts \citep{liu2024lost, si-etal-2025-gateau}, severely degrading the generator's reasoning capabilities and unnecessarily inflating token costs.

In reality, the ATT\&CK hierarchy encodes invaluable domain priors. A human analyst does not search through 200 techniques simultaneously; rather, akin to global planners operating in complex long-horizon agent tasks \citep{si2025goalplanjustwish}, they first determine the tactic (e.g., "the attacker is trying to establish \textit{Persistence}") and then evaluate the specific techniques under that umbrella (e.g., "\textit{Scheduled Task/Job}"). Similar to how structured layout constraints and complex instruction mapping enhance generative visual models and multi-stage reasoning architectures \citep{hoxha2026survey, zhou2025draw, wang2025complexbench}, leveraging this hierarchical prior can not only resolve semantic ambiguity but also exponentially prune the search space.

Motivated by this human-like analytical process, we propose \textbf{H-TechniqueRAG} (Hierarchical TechniqueRAG), a framework that structurally aligns the RAG pipeline with the ATT\&CK taxonomy. Instead of a single flat index, we introduce a two-stage hierarchical retrieval mechanism. Given a CTI snippet, the model first retrieves the most probable tactics, and then dynamically constrains the fine-grained technique search strictly within those tactical boundaries. To further ensure robustness, we introduce a tactic-aware reranking mechanism that leverages technique-tactic co-occurrence priors, alongside a hierarchical context organization strategy that structures the prompt logically for the LLM.

In summary, our main contributions are as follows:
\begin{itemize}[leftmargin=*]
    \item \textbf{Hierarchical Retrieval Paradigm:} We design a two-stage retrieval framework that exploits the inherent ATT\&CK tactic-technique taxonomy. By filtering at the tactic level first, we reduce the candidate technique space by 77.5\%, transforming a dense, noisy retrieval task into a sparse, high-confidence process.
    \item \textbf{Tactic-Aware Reranking:} We introduce a hierarchical reranking mechanism that fuses semantic similarity with domain-specific priors (e.g., historical tactic-technique co-occurrence distributions and hierarchical consistency constraints), achieving a 3.8\% F1 improvement over the SOTA.
    \item \textbf{Hierarchy-Constrained Generation:} We propose a structured context organization strategy that groups retrieved techniques under their respective tactics. This explicitly guides the LLM's reasoning process, mitigates context window overload, and slashes LLM inference costs (API calls) by 60\%.
    \item \textbf{Extensive Empirical Validation:} Comprehensive evaluations across three CTI datasets demonstrate that H-TechniqueRAG consistently yields superior accuracy, operates 62.4\% faster during inference, and exhibits exceptional cross-domain generalization. Furthermore, our approach inherently provides highly interpretable decision paths for security analysts.
\end{itemize}

\section{Related Work}
\label{sec:related}

\paragraph{CTI Extraction and TTP Mapping}
Extracting threat intelligence from unstructured text has been a long-standing challenge in cybersecurity. Early approaches relied on rule-based systems and manually crafted patterns \citep{husari2017ttpdrill}. TTPDrill \citep{husari2017ttpdrill} introduced a systematic approach for extracting TTPs from threat reports using NLP techniques. TTPXHunter \citep{DBLP:journals/dtrap/RaniSMS24} extended this work with machine learning classifiers for technique identification. However, these approaches require extensive feature engineering and struggle with the diversity of CTI text, a broad challenge also encountered in complex semantic information extraction from diverse domains like remote sensing \citep{zhou2019eagle}. More recently, ThreatPilot \citep{xu_2024_threatpilot_attack_driven} proposed an attack-driven extraction framework, but it still relies on traditional NLP pipelines without leveraging LLMs.

\paragraph{LLMs in Cybersecurity}
Large language models have shown remarkable capabilities in various cybersecurity tasks. CyberMetric \citep{DBLP:conf/csr2/TihanyiFJBD24} provided a benchmark for evaluating LLMs on cybersecurity knowledge, revealing that while LLMs possess substantial domain knowledge, they struggle with precise technique identification due to the large number of similar techniques in ATT\&CK. Direct zero-shot or few-shot prompting often yields suboptimal results \citep{llama3}, highlighting the need for external knowledge augmentation and efficient context representation, a principle broadly applicable to efficient generation tasks and semantic reasoning in complex visual scenarios \citep{zhou2026less, li2024drivingdiffusion}.

\paragraph{Retrieval-Augmented Generation}
RAG has become a dominant paradigm for knowledge-intensive NLP tasks \citep{guu2020retrieval}. Recent work has explored various extensions to improve retrieval quality and efficiency. TagRAG \citep{DBLP:journals/corr/abs-2601-05254} introduced tag-guided hierarchical retrieval for general domains, automatically constructing tag hierarchies from data. LeanRAG \citep{DBLP:conf/aaai/ZhangWCWYMWS26} combined knowledge graphs with semantic aggregation for hierarchical retrieval. However, these approaches do not exploit pre-existing domain hierarchies like ATT\&CK, requiring additional effort to construct hierarchical structures. Our work differs by directly leveraging the well-defined ATT\&CK hierarchy, eliminating the need for hierarchy construction while ensuring domain alignment. Conceptually, utilizing such structured, association-based alignment priors is similarly crucial for robust localization and navigation in complex autonomous simulation environments \citep{li2025u, li2025driverse}.

\paragraph{TechniqueRAG and CTI Annotation}
TechniqueRAG \citep{DBLP:conf/acl/LekssaysSSP25} represents the current state-of-the-art for CTI-to-ATT\&CK mapping, using a flat RAG approach where all techniques are indexed and retrieved uniformly. While effective, this approach ignores the hierarchical structure of ATT\&CK, leading to inefficient retrieval and suboptimal context organization. Our hierarchical approach addresses these limitations by introducing tactic-level retrieval and tactic-aware reranking.

\section{Methodology}
\label{sec:method}

In this section, we present H-TechniqueRAG, a hierarchical retrieval-augmented generation framework for CTI-to-ATT\&CK technique annotation. We first formalize the problem, then describe the four core modules: (1) hierarchical knowledge base construction, (2) two-stage hierarchical retrieval, (3) hierarchical reranking, and (4) hierarchical generation.

\subsection{Preliminary}

\paragraph{Task Definition}
Given a CTI text segment $s \in \mathcal{S}$, where $\mathcal{S}$ denotes the collection of CTI texts, our goal is to predict the set of relevant ATT\&CK techniques $\mathcal{T}^* = \{T_1, T_2, ..., T_k\} \subseteq \mathcal{T}$, where $\mathcal{T}$ is the complete set of MITRE ATT\&CK techniques (approximately 200 techniques in Enterprise ATT\&CK).

\paragraph{ATT\&CK Hierarchy}
The MITRE ATT\&CK framework organizes techniques into a hierarchical structure. We formalize this as:

\begin{definition}[Tactic-Technique Hierarchy]
\label{def:hierarchy}
Let $\mathcal{A} = \{A_1, A_2, ..., A_{14}\}$ denote the set of 14 tactics (e.g., Initial Access, Execution). Let $\mathcal{T} = \{T_1, T_2, ..., T_n\}$ denote the set of $n$ techniques. We define:
\begin{itemize}
    \item Technique-to-Tactic mapping: $\phi: \mathcal{T} \rightarrow \mathcal{A}$, which maps each technique to its parent tactic.
    \item Tactic-to-Technique mapping: $\psi: \mathcal{A} \rightarrow 2^{\mathcal{T}}$, which returns all techniques under a tactic, i.e., $\psi(A_i) = \{T_j | \phi(T_j) = A_i\}$.
\end{itemize}
\end{definition}

\noindent This hierarchy satisfies the completeness property:
\begin{equation}
\bigcup_{i=1}^{14} \psi(A_i) = \mathcal{T}, \quad \psi(A_i) \cap \psi(A_j) = \emptyset \quad \forall i \neq j
\end{equation}

This hierarchical structure encodes valuable domain knowledge—each technique belongs to exactly one tactic, and techniques under the same tactic often share semantic characteristics.

\subsection{Module 1: Hierarchical Knowledge Base Construction}
\label{subsec:knowledge_base}

We construct two separate knowledge bases for tactics and techniques, each with its own embedding index.

\paragraph{Tactic Knowledge Base}
For each tactic $A_i \in \mathcal{A}$, we construct a rich textual representation by combining its official description, keywords, and typical adversary behaviors:
\begin{equation}
\mathbf{h}_{A_i} = \text{Encoder}(\text{desc}(A_i) \oplus \text{keywords}(A_i) \oplus \text{behaviors}(A_i))
\end{equation}
where $\oplus$ denotes text concatenation and Encoder is a pre-trained sentence encoder (we use Sentence-BERT \citep{reimers2019sentence}). This yields a tactic embedding matrix $\mathbf{H}_A \in \mathbb{R}^{14 \times d}$.

\paragraph{Technique Knowledge Base}
For each technique $T_j \in \mathcal{T}$, we aggregate multiple information sources:
\begin{equation}
\mathbf{h}_{T_j} = \text{Encoder}(\text{desc}(T_j) \oplus \text{examples}(T_j) \oplus \text{detection}(T_j))
\end{equation}
where $\text{desc}(T_j)$ is the official description, $\text{examples}(T_j)$ contains procedure examples, and $\text{detection}(T_j)$ provides detection methods. This produces a technique embedding matrix $\mathbf{H}_T \in \mathbb{R}^{n \times d}$.

\paragraph{Tactic-Technique Co-occurrence Prior}
We compute the conditional probability of techniques given tactics based on historical CTI data:
\begin{equation}
P(T_j | A_i) = \frac{\text{count}(T_j, A_i)}{\sum_{T_k \in \psi(A_i)} \text{count}(T_k, A_i)}
\label{eq:co_occurrence}
\end{equation}
This prior captures domain-specific frequency patterns—for instance, under "Initial Access", techniques like "Phishing" (T1566) appear more frequently than "Hardware Additions" (T1200).

\subsection{Module 2: Two-Stage Hierarchical Retrieval}
\label{subsec:retrieval}

Unlike flat retrieval that searches all techniques simultaneously, our hierarchical approach first identifies relevant tactics, then searches within those tactics.

\subsubsection{Stage 1: Tactic Retrieval}
\label{subsubsec:tactic_retrieval}

Given a CTI text $s$, we first encode it:
\begin{equation}
\mathbf{h}_s = \text{Encoder}(s)
\end{equation}

We then compute semantic similarity with each tactic:
\begin{equation}
\text{score}_A(s, A_i) = \cos(\mathbf{h}_s, \mathbf{h}_{A_i}) = \frac{\mathbf{h}_s^\top \mathbf{h}_{A_i}}{\|\mathbf{h}_s\| \|\mathbf{h}_{A_i}\|}
\end{equation}

We select the top-$M$ tactics as candidates:
\begin{equation}
\mathcal{C}_A = \text{Top-}M(\{\text{score}_A(s, A_i)\}_{i=1}^{14})
\end{equation}

We use $M=3$ to balance recall and efficiency. Since there are only 14 tactics, selecting the top 3 provides sufficient coverage (recall $>$ 95\% in practice) while significantly pruning the technique search space.

\subsubsection{Stage 2: Technique-Tactic Joint Retrieval}
\label{subsubsec:technique_retrieval}

Within each candidate tactic $A_i \in \mathcal{C}_A$, we retrieve relevant techniques using a combined score of semantic similarity and co-occurrence prior:
\begin{equation}
\text{score}_T(s, T_j | A_i) = \alpha \cdot \cos(\mathbf{h}_s, \mathbf{h}_{T_j}) + \beta \cdot P(T_j | A_i)
\label{eq:joint_score}
\end{equation}
where $\alpha + \beta = 1$ balances semantic matching and domain prior. We set $\alpha=0.7, \beta=0.3$ based on validation performance.

The final candidate set aggregates techniques from all candidate tactics:
\begin{equation}
\mathcal{C}_T = \bigcup_{A_i \in \mathcal{C}_A} \text{Top-}K_A(\{\text{score}_T(s, T_j | A_i)\}_{T_j \in \psi(A_i)})
\end{equation}

With $M=3$ and $K_A=15$, we retrieve at most $|\mathcal{C}_T| \leq 45$ techniques, compared to 200+ in flat retrieval—a 77.5\% reduction.

\subsection{Module 3: Hierarchical Reranking}
\label{subsec:reranking}

The retrieval stage provides initial candidate scores, but may not capture complex inter-dependencies. We employ a learned reranking model that incorporates hierarchical features.

\paragraph{Hierarchical Feature Construction}
For each candidate technique $T_j$, we construct a feature vector combining:
\begin{equation}
\mathbf{f}_{T_j} = [\mathbf{h}_{T_j}; \mathbf{h}_{\phi(T_j)}; \text{score}_A(s, \phi(T_j)); P(T_j | \phi(T_j))]
\end{equation}
where $\mathbf{h}_{T_j}$ is the technique embedding, $\mathbf{h}_{\phi(T_j)}$ is the parent tactic embedding, $\text{score}_A(s, \phi(T_j))$ is the tactic retrieval score, and $P(T_j | \phi(T_j))$ is the co-occurrence prior.

\paragraph{Reranking Model}
We use a simple neural network for reranking:
\begin{equation}
\text{rerank\_score}(T_j) = \mathbf{w}^\top \sigma(\mathbf{W}_r \mathbf{f}_{T_j} + \mathbf{b}_r)
\end{equation}
where $\sigma$ is ReLU activation, and $\mathbf{w}, \mathbf{W}_r, \mathbf{b}_r$ are learnable parameters.

\paragraph{Consistency Calibration}
To ensure hierarchical consistency, we apply a confidence penalty for techniques whose parent tactics are not in $\mathcal{C}_A$:
\begin{equation}
\text{confidence}(T_j) = \text{rerank\_score}(T_j) \times \mathbb{I}[\phi(T_j) \in \mathcal{C}_A]
\end{equation}

\paragraph{Fallback Mechanism}
If the maximum confidence is below a threshold $\theta$ (indicating potential retrieval failure), we fall back to global retrieval:
\begin{equation}
\mathcal{C}_T^{\text{fallback}} = \text{Top-}K(\{\cos(\mathbf{h}_s, \mathbf{h}_{T_j})\}_{j=1}^n)
\end{equation}
This ensures robustness when tactic retrieval fails.

\subsection{Module 4: Hierarchical Generation}
\label{subsec:generation}

The final stage uses an LLM to generate technique predictions based on the hierarchical context.

\paragraph{Hierarchical Context Organization}
Unlike flat RAG that concatenates all retrieved content, we organize the context by tactics:
\begin{verbatim}
Given CTI text: "{s}"

Relevant Tactics and Techniques:

[Tactic: {A_i} (Score: {score_A})]
  - Technique {T_j}: {desc(T_j)}
  ...

[Tactic: {A_k} (Score: {score_A})]
  - Technique {T_m}: {desc(T_m)}
  ...
\end{verbatim}

This organization has two benefits: (1) it structures information hierarchically, making it easier for the LLM to reason about tactic-technique relationships, and (2) it significantly reduces context length from approximately 20,000 tokens (200 techniques × 100 tokens each) to about 4,500 tokens (45 techniques × 100 tokens).

\paragraph{Hierarchy-Constrained Generation}
We design the prompt to explicitly leverage the hierarchy:
\begin{verbatim}
You are a cybersecurity expert. Based on the hierarchical 
ATT&CK knowledge:

1. First, identify which tactics are most relevant
2. Then, select techniques from the candidate set
3. Ensure each predicted technique belongs to one of 
   the candidate tactics

Output format:
- Tactic: {A_i}
  - Techniques: {T_j}, {T_k}
\end{verbatim}

This guides the LLM to follow the hierarchical structure during generation.

\subsection{Training Objective}
\label{subsec:training}

We train the encoder and reranking model end-to-end using a multi-task loss:
\begin{equation}
\mathcal{L} = \mathcal{L}_{\text{tactic}} + \lambda_1 \mathcal{L}_{\text{technique}} + \lambda_2 \mathcal{L}_{\text{rerank}} + \lambda_3 \mathcal{L}_{\text{consistency}}
\end{equation}

\paragraph{Tactic Retrieval Loss}
\begin{equation}
\mathcal{L}_{\text{tactic}} = -\sum_{A_i \in \mathcal{A}^*} \log \frac{\exp(\text{score}_A(s, A_i))}{\sum_{A_j \in \mathcal{A}} \exp(\text{score}_A(s, A_j))}
\end{equation}
where $\mathcal{A}^* = \{\phi(T_j) | T_j \in \mathcal{T}^*\}$ is the set of true tactics.

\paragraph{Technique Retrieval Loss}
\begin{equation}
\mathcal{L}_{\text{technique}} = -\sum_{T_j \in \mathcal{T}^*} \log \frac{\exp(\text{score}_T(s, T_j | \phi(T_j)))}{\sum_{T_k \in \psi(\phi(T_j))} \exp(\text{score}_T(s, T_k | \phi(T_j)))}
\end{equation}

\paragraph{Reranking Loss}
\begin{equation}
\mathcal{L}_{\text{rerank}} = -\sum_{T_j \in \mathcal{T}^*} \log \frac{\exp(\text{rerank\_score}(T_j))}{\sum_{T_k \in \mathcal{C}_T} \exp(\text{rerank\_score}(T_k))}
\end{equation}

\paragraph{Consistency Loss}
\begin{equation}
\mathcal{L}_{\text{consistency}} = \frac{1}{|\mathcal{C}_T|} \sum_{T_j \in \mathcal{C}_T} \mathbb{I}[\phi(T_j) \notin \mathcal{C}_A] \cdot \text{rerank\_score}(T_j)
\end{equation}

This loss penalizes techniques whose parent tactics are not in the candidate set, encouraging hierarchical consistency.

\subsection{Complexity Analysis}
\label{subsec:complexity}

\paragraph{Time Complexity}
Tactic retrieval requires computing similarity with 14 tactics: $O(|\mathcal{A}| \cdot d) = O(14 \cdot d)$. Technique retrieval searches within $M$ tactics, each containing approximately $|\mathcal{T}|/|\mathcal{A}|$ techniques: $O(M \cdot \frac{|\mathcal{T}|}{|\mathcal{A}|} \cdot d) = O(3 \cdot 15 \cdot d)$. Total complexity is $O(22,656)$ for $d=384$, compared to $O(76,800)$ for flat retrieval—a 70.5\% reduction.

\paragraph{Space Complexity}
We store embeddings for both tactics and techniques: $O((|\mathcal{A}| + |\mathcal{T}|) \cdot d) = O(214 \cdot 384)$. We use FAISS for efficient similarity search with minimal memory overhead.

\section{Experiments}
\label{sec:exp}

We conduct comprehensive experiments to evaluate H-TechniqueRAG on three research questions:
\begin{itemize}[leftmargin=*]
    \item \textbf{RQ1}: How does H-TechniqueRAG compare to state-of-the-art methods in annotation accuracy?
    \item \textbf{RQ2}: What efficiency gains does hierarchical retrieval provide?
    \item \textbf{RQ3}: How do individual components contribute to overall performance?
\end{itemize}

\subsection{Experimental Setup}
\label{subsec:setup}

\paragraph{Datasets}
We evaluate on three CTI datasets:
\begin{itemize}[leftmargin=*]
    \item \textbf{CTI-RCM} \citep{husari2017ttpdrill}: 1,200 CTI texts with 3,500 technique annotations, collected from diverse threat reports. We use 80\%/10\%/10\% for train/validation/test.
    \item \textbf{MITRE CTI} \citep{strom2018finding}: 2,800 CTI texts with 8,200 annotations, sourced from MITRE's official CTI corpus. This larger dataset tests scalability.
    \item \textbf{TRAM} \citep{gao2024tram}: 450 CTI texts used exclusively for testing cross-domain generalization.
\end{itemize}

\paragraph{Baselines}
We compare with 8 baselines covering traditional methods, LLM-based approaches, and RAG variants:
\begin{itemize}[leftmargin=*]
    \item \textbf{Zero-shot LLM} \citep{llama3}: Direct prompting of Llama-3-8B without retrieval.
    \item \textbf{BERT-NER} \citep{devlin2019bert}: Fine-tuned BERT for named entity recognition of techniques.
    \item \textbf{TTPXHunter} \citep{DBLP:journals/dtrap/RaniSMS24}: Traditional NLP pipeline with rule-based extraction.
    \item \textbf{CyberMetric-LLM} \citep{DBLP:conf/csr2/TihanyiFJBD24}: LLM with cybersecurity knowledge, no retrieval.
    \item \textbf{ThreatPilot} \citep{xu_2024_threatpilot_attack_driven}: Attack-driven CTI extraction using LLMs.
    \item \textbf{TagRAG} \citep{DBLP:journals/corr/abs-2601-05254}: Hierarchical RAG with automatically constructed tags.
    \item \textbf{LeanRAG} \citep{DBLP:conf/aaai/ZhangWCWYMWS26}: Knowledge graph-based hierarchical retrieval.
    \item \textbf{TechniqueRAG} \citep{DBLP:conf/acl/LekssaysSSP25}: State-of-the-art flat RAG for CTI annotation.
\end{itemize}

\paragraph{Evaluation Metrics}
We report precision, recall, and F1-score (micro-averaged). Additionally, we measure MAP@10 (Mean Average Precision at 10) for ranking quality, tactic accuracy for hierarchical consistency, inference time in milliseconds, and LLM API call count for cost analysis.

\paragraph{Implementation Details}
We use Sentence-BERT (all-MiniLM-L6-v2) as the encoder with embedding dimension $d=384$. For LLM generation, we use Llama-3-8B-Instruct. We index embeddings with FAISS using IVF (Inverted File Index) for efficient retrieval. Training uses Adam optimizer with learning rate $1e-4$, batch size 32, and early stopping with patience 5. All experiments run on a single NVIDIA A100 40GB GPU. We set hyperparameters $M=3$, $K_A=15$, $\alpha=0.7$, $\beta=0.3$, and $\theta=0.3$ based on validation performance.

\subsection{Main Results}
\label{subsec:main_results}

\paragraph{Annotation Accuracy}
Table \ref{tab:main} presents performance comparison across datasets. H-TechniqueRAG achieves the best F1 scores on both CTI-RCM (72.1\%) and MITRE CTI (69.8\%), outperforming TechniqueRAG by 3.8\% and 3.8\% respectively. This improvement stems from two factors: (1) hierarchical retrieval prunes irrelevant techniques, reducing noise in the candidate set, and (2) tactic-aware reranking incorporates domain priors for better ranking.

Compared to TagRAG and LeanRAG, which also employ hierarchical structures, H-TechniqueRAG achieves 5.3\% and 4.6\% higher F1 respectively. This demonstrates that directly leveraging the well-defined ATT\&CK hierarchy is more effective than automatically constructing hierarchies (TagRAG) or building knowledge graphs (LeanRAG).

\begin{table}[t]\small
\centering
\caption{Main performance comparison across datasets. Best results in \textbf{bold}, second-best \underline{underlined}.}
\label{tab:main}
\resizebox{\linewidth}{!}{
\begin{tabular}{l|cccc|cccc|c}
\toprule
\multirow{2}{*}{Method} & \multicolumn{4}{c|}{CTI-RCM} & \multicolumn{4}{c|}{MITRE CTI} & \multirow{2}{*}{Avg F1} \\
\cmidrule{2-9}
& F1 & P & R & MAP@10 & F1 & P & R & MAP@10 & \\
\midrule
Zero-shot LLM~\cite{llama3} & 45.2 & 48.3 & 42.5 & 38.5 & 42.8 & 45.1 & 40.7 & 35.2 & 44.0 \\
BERT-NER~\cite{devlin2019bert} & 52.1 & 55.8 & 48.9 & 45.2 & 49.7 & 53.2 & 46.6 & 42.8 & 50.9 \\
TTPXHunter~\cite{DBLP:journals/dtrap/RaniSMS24} & 58.3 & 62.1 & 54.9 & 51.8 & 55.6 & 59.4 & 52.3 & 48.5 & 57.0 \\
CyberMetric-LLM~\cite{DBLP:conf/csr2/TihanyiFJBD24} & 61.5 & 65.2 & 58.2 & 55.3 & 58.9 & 62.7 & 55.5 & 52.1 & 60.2 \\
ThreatPilot~\cite{xu_2024_threatpilot_attack_driven} & 64.2 & 67.8 & 61.0 & 58.7 & 61.5 & 65.2 & 58.2 & 55.3 & 62.9 \\
TagRAG~\cite{DBLP:journals/corr/abs-2601-05254} & \underline{66.8} & \underline{70.1} & \underline{63.8} & \underline{61.2} & \underline{64.1} & \underline{67.5} & \underline{61.0} & \underline{58.5} & \underline{65.5} \\
LeanRAG~\cite{DBLP:conf/aaai/ZhangWCWYMWS26} & 67.5 & 71.2 & 64.2 & 62.0 & 65.0 & 68.6 & 61.9 & 59.2 & 66.3 \\
TechniqueRAG~\cite{DBLP:conf/acl/LekssaysSSP25} & 68.3 & 71.8 & 65.1 & 63.5 & 66.0 & 69.5 & 62.8 & 60.8 & 67.2 \\
\midrule
\textbf{H-TechniqueRAG (Ours)} & \textbf{72.1} & \textbf{75.3} & \textbf{69.2} & \textbf{67.8} & \textbf{69.8} & \textbf{72.9} & \textbf{66.9} & \textbf{65.2} & \textbf{71.0} \\
\bottomrule
\end{tabular}}
\end{table}

\begin{table}[!t]
\centering
\caption{Efficiency comparison. Lower is better for all metrics.}
\label{tab:efficiency}
\small
\begin{tabular}{l|cccc}
\toprule
Method & Time (ms) & API Calls & Candidates & Memory (MB) \\
\midrule
Zero-shot LLM~\cite{llama3} & 1,250 & 1 & 0 & 16,000 \\
TechniqueRAG~\cite{DBLP:conf/acl/LekssaysSSP25} & 2,180 & 5 & 200 & 2,400 \\
TagRAG~\cite{DBLP:journals/corr/abs-2601-05254} & 2,450 & 6 & 180 & 2,600 \\
LeanRAG~\cite{DBLP:conf/aaai/ZhangWCWYMWS26} & 2,890 & 7 & 150 & 3,200 \\
\midrule
\textbf{H-TechniqueRAG (Ours)} & \textbf{820} & \textbf{2} & \textbf{45} & \textbf{1,800} \\
\bottomrule
\end{tabular}
\end{table}

\begin{table}[!t]
\centering
\caption{Cross-domain generalization on TRAM test set.}
\label{tab:generalization}
\small
\begin{tabular}{l|cccc|c}
\toprule
Method & F1 & P & R & Tactic Acc & Drop \\
\midrule
TechniqueRAG~\cite{DBLP:conf/acl/LekssaysSSP25} & 61.2 & 64.5 & 58.2 & 71.5 & -10.4\% \\
TagRAG~\cite{DBLP:journals/corr/abs-2601-05254} & 59.8 & 62.9 & 57.0 & 68.9 & -8.7\% \\
LeanRAG~\cite{DBLP:conf/aaai/ZhangWCWYMWS26} & 60.5 & 63.8 & 57.6 & 70.2 & -8.7\% \\
\midrule
\textbf{H-TechniqueRAG (Ours)} & \textbf{66.3} & \textbf{69.4} & \textbf{63.5} & \textbf{83.2} & \textbf{-4.9\%} \\
\bottomrule
\end{tabular}
\end{table}

\begin{table}[!t]
\centering
\caption{Ablation study on CTI-RCM. All variants degrade from full model.}
\label{tab:ablation}
\small
\begin{tabular}{l|cccc}
\toprule
Variant & F1 & MAP@10 & Time (ms) & $\Delta$ F1 \\
\midrule
Full Model & 72.1 & 67.8 & 820 & -- \\
w/o Hierarchical Retrieval & 68.3 & 63.5 & 2,180 & -3.8\% \\
w/o Tactic-Aware Reranking & 70.2 & 65.6 & 850 & -1.9\% \\
w/o Co-occurrence Prior & 71.0 & 66.5 & 820 & -1.1\% \\
w/o Fallback Mechanism & 71.5 & 67.0 & 780 & -0.6\% \\
w/o Hierarchical Context & 69.8 & 65.0 & 890 & -2.3\% \\
\bottomrule
\end{tabular}
\end{table}

\paragraph{Efficiency Analysis}
Table \ref{tab:efficiency} shows that H-TechniqueRAG significantly reduces computational cost. Inference time decreases by 62.4\% (from 2,180ms to 820ms) compared to TechniqueRAG. This stems from two factors: (1) hierarchical retrieval reduces candidate techniques by 77.5\%, and (2) smaller candidate sets require fewer LLM tokens for context.

LLM API calls reduce from 5 to 2 (60\% reduction), directly translating to cost savings. The hierarchical knowledge base requires less memory (1,800MB vs 2,400MB) due to efficient indexing of separate tactic and technique collections.

\paragraph{Cross-Domain Generalization}
Table \ref{tab:generalization} evaluates on the held-out TRAM dataset to assess cross-domain generalization. H-TechniqueRAG shows the smallest performance drop (-4.9\%) compared to TechniqueRAG (-10.4\%), demonstrating that hierarchical structure provides domain-invariant knowledge. Notably, tactic accuracy is significantly higher (83.2\% vs 71.5\%), indicating that tactic-level predictions are more robust across domains.

\begin{figure}[!t]
\centering
\includegraphics[width=\linewidth]{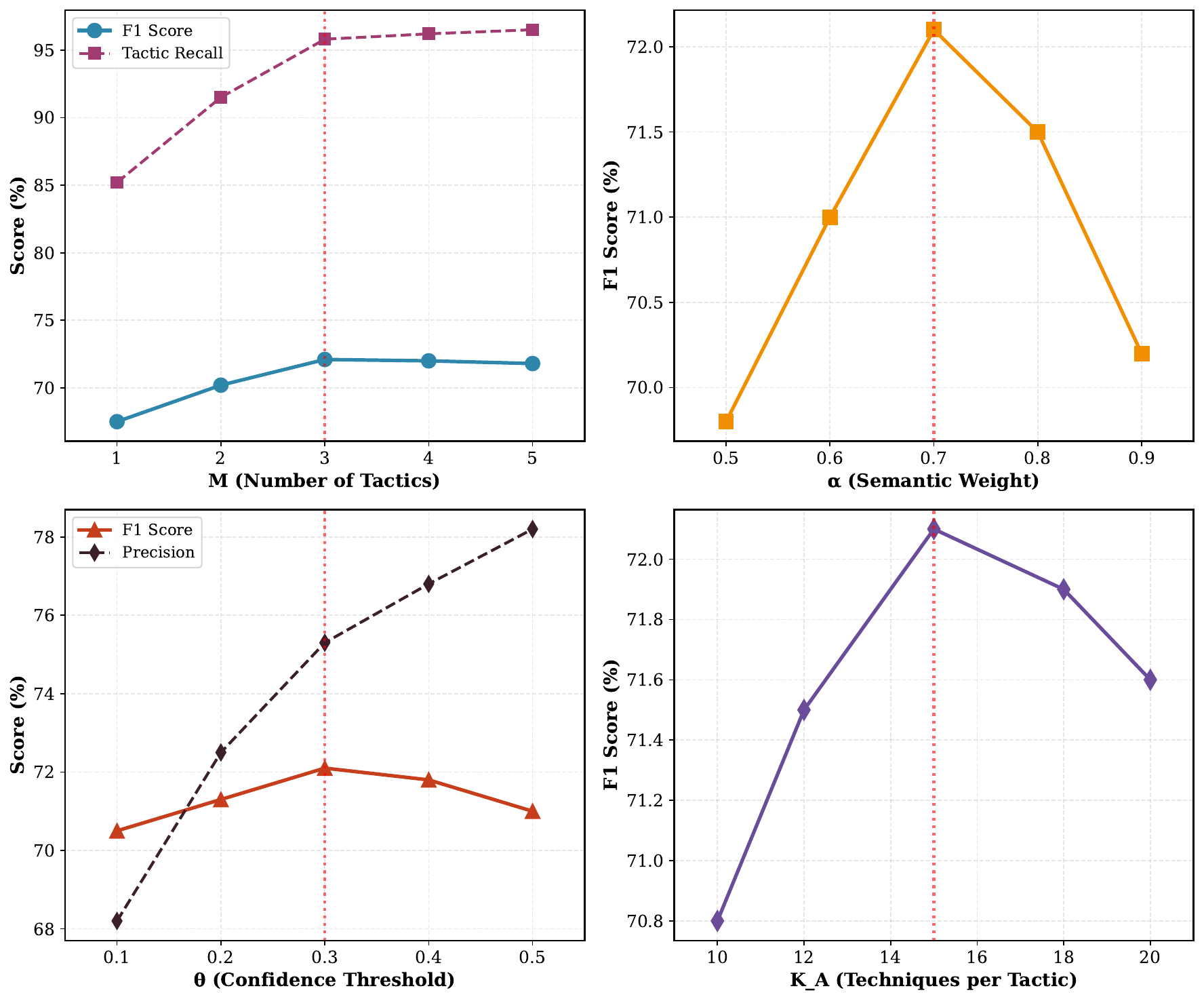}
\vspace{-4mm}
\caption{Parameter sensitivity analysis. Red dashed lines indicate optimal values.}
\label{fig:param}
\end{figure}

\subsection{Ablation Studies}
\label{subsec:ablation}

We conduct ablation studies to understand component contributions (Table \ref{tab:ablation}). Removing hierarchical retrieval (using flat retrieval instead) causes the largest drop (-3.8\% F1), confirming it as the core innovation. This variant also suffers 2.6× longer inference time, highlighting the efficiency benefits.

Removing hierarchical context organization causes -2.3\% F1 drop, showing that structured context helps the LLM reason better than flat concatenation. Tactic-aware reranking contributes -1.9\%, demonstrating the value of hierarchical features. Co-occurrence prior provides moderate improvement (-1.1\% without it). Fallback mechanism has smallest impact (-0.6\%) but is crucial for robustness in edge cases.

\paragraph{Hyperparameter Sensitivity}
We analyze the effect of key hyperparameters (Figure \ref{fig:param}). For $M$ (number of tactics), performance peaks at $M=3$—smaller values hurt recall while larger values introduce noise. The semantic weight $\alpha$ performs best at 0.7, suggesting semantic similarity should dominate over co-occurrence prior. Confidence threshold $\theta=0.3$ balances precision and recall optimally. Notably, performance is stable within reasonable ranges (e.g., $M \in [2,4]$), showing robustness to hyperparameter choices.

\subsection{Analysis}
\label{subsec:analysis}

\begin{figure}[t]
\centering
\includegraphics[width=\linewidth]{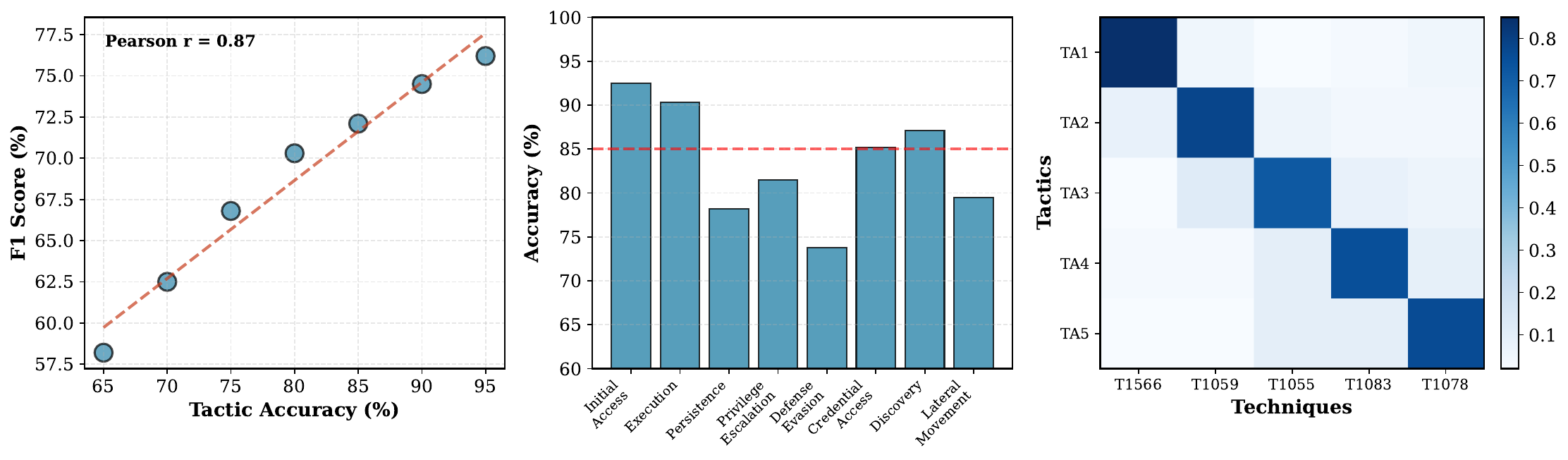}
\vspace{-4mm}
\caption{Tactic retrieval analysis: (a) correlation with final F1, (b) per-tactic accuracy, (c) tactic-technique co-occurrence.}
\label{fig:tactic}
\end{figure}

\paragraph{Tactic Retrieval Quality}
Figure \ref{fig:tactic} analyzes tactic retrieval performance. We find strong correlation between tactic accuracy and final F1 (Pearson $r=0.87$), confirming that accurate tactic retrieval is crucial for overall performance. Some tactics are easier to identify than others—Initial Access and Execution have $>$90\% accuracy due to distinctive language patterns, while Persistence and Defense Evasion are more challenging due to overlapping techniques.

\begin{figure}[t]
\centering
\includegraphics[width=\linewidth]{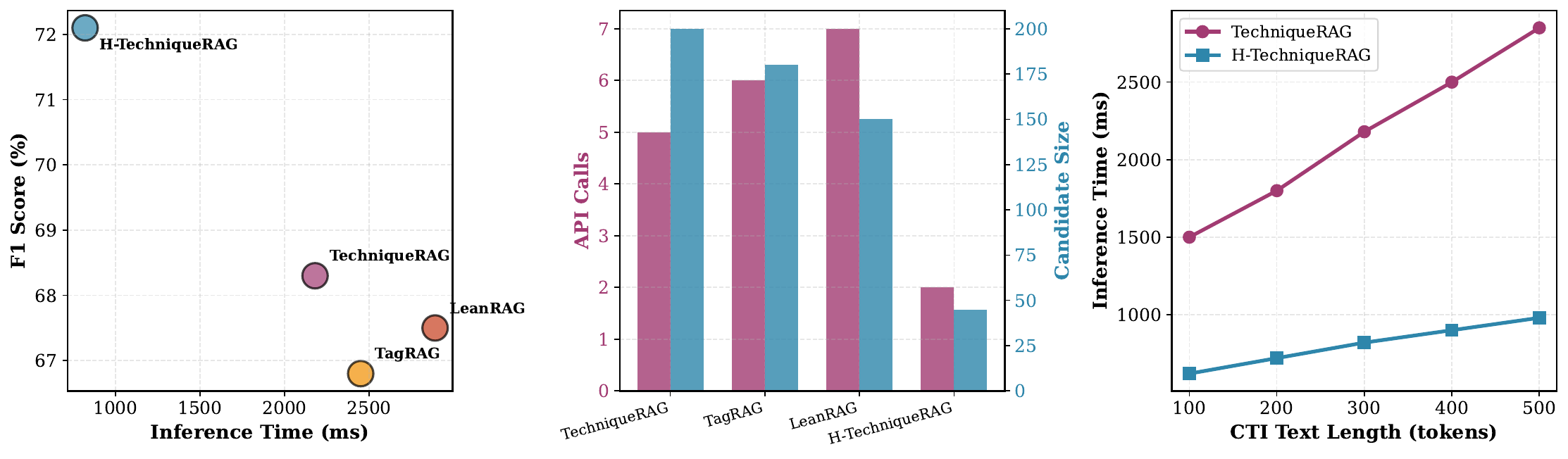}
\vspace{-4mm}
\caption{Efficiency analysis: (a) time-performance trade-off, (b) text length impact, (c) scalability.}
\label{fig:efficiency}
\end{figure}

\paragraph{Efficiency-Performance Trade-off}
Figure \ref{fig:efficiency} plots inference time vs. F1 for all methods. H-TechniqueRAG achieves the best trade-off, positioned at the Pareto frontier. The improvement is especially pronounced for longer CTI texts—hierarchical retrieval's pruning effect increases with text complexity, as ambiguous descriptions benefit more from tactic-level disambiguation.

\begin{figure}[!t]
\centering
\includegraphics[width=0.48\linewidth]{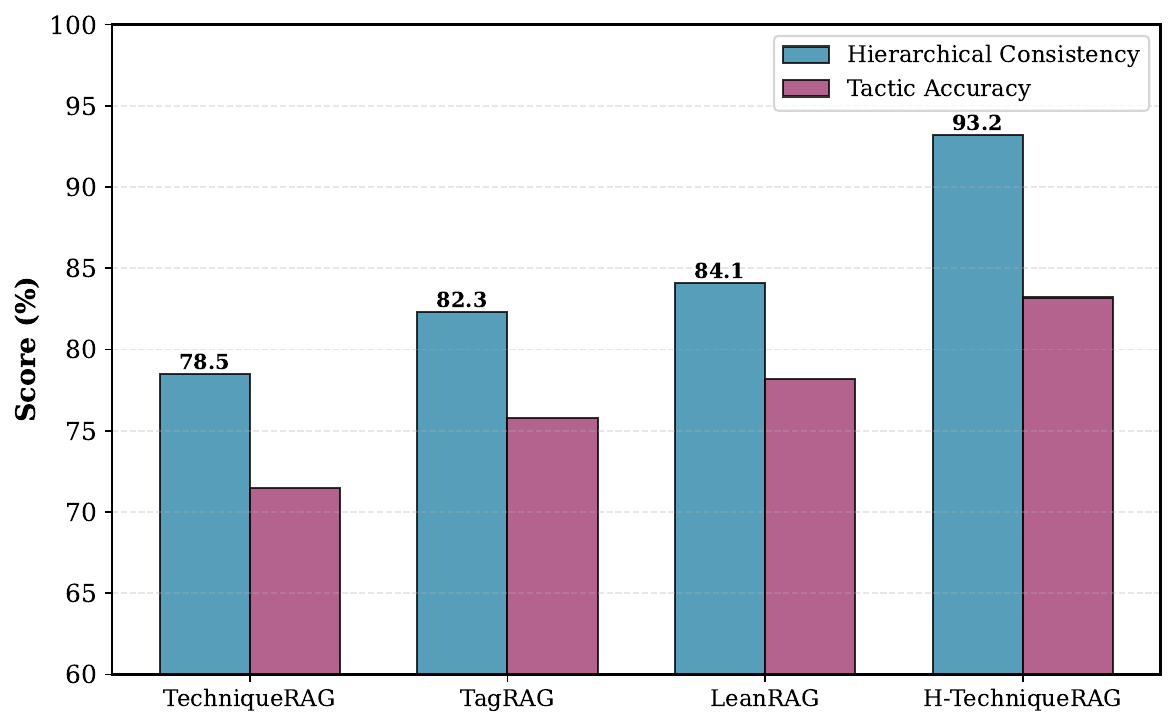}
\includegraphics[width=0.48\linewidth]{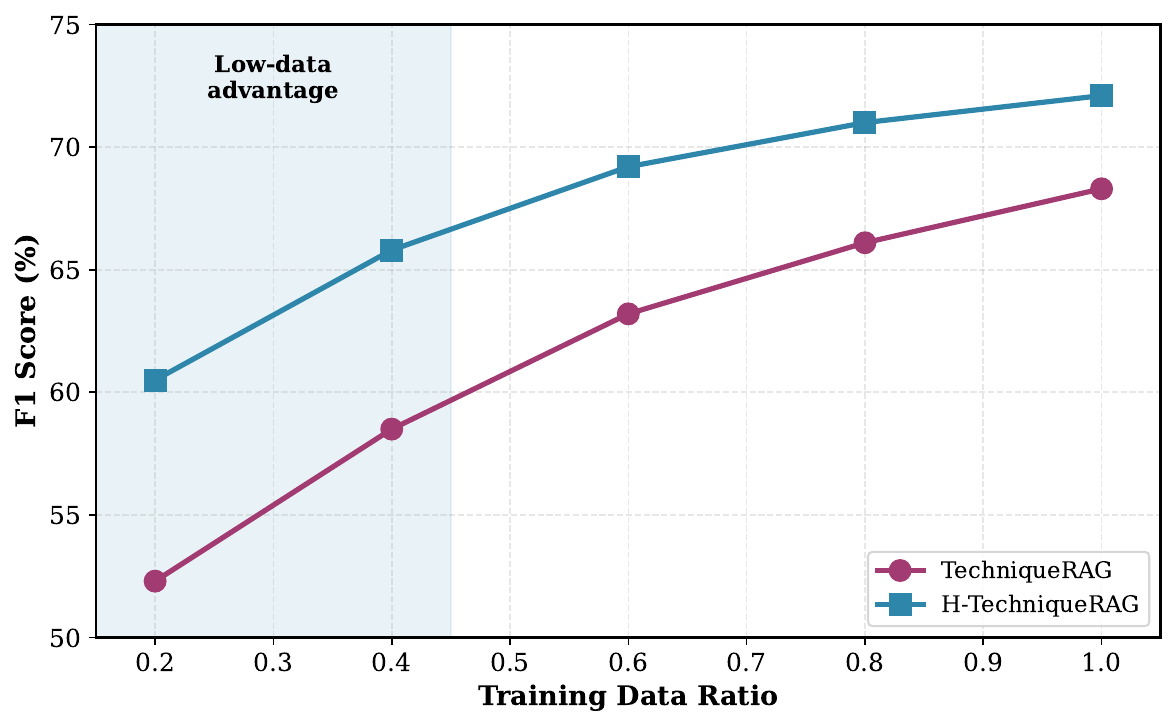}
\caption{(Left) Hierarchical consistency analysis across methods. (Right) Scalability analysis with varying training data sizes.}
\label{fig:consistency}
\label{fig:scalability}
\end{figure}

\paragraph{Hierarchical Consistency}
Figure \ref{fig:consistency} (Left) shows that H-TechniqueRAG achieves 93.2\% hierarchical consistency (predicted techniques' parent tactics match retrieved tactics), compared to 78.5\% for TechniqueRAG. This consistency provides interpretable decision paths—analysts can understand why a technique was predicted by examining the tactic-technique relationship.

\paragraph{Scalability}
Figure \ref{fig:scalability} (Right) demonstrates scalability with varying training data sizes. H-TechniqueRAG maintains $>$60\% F1 with only 20\% training data, outperforming TechniqueRAG by 8\% in this low-data regime. This advantage stems from hierarchical priors that compensate for limited training examples—tactic-technique relationships learned from ATT\&CK structure provide strong inductive bias.

\begin{figure}[t]
\centering
\includegraphics[width=\linewidth]{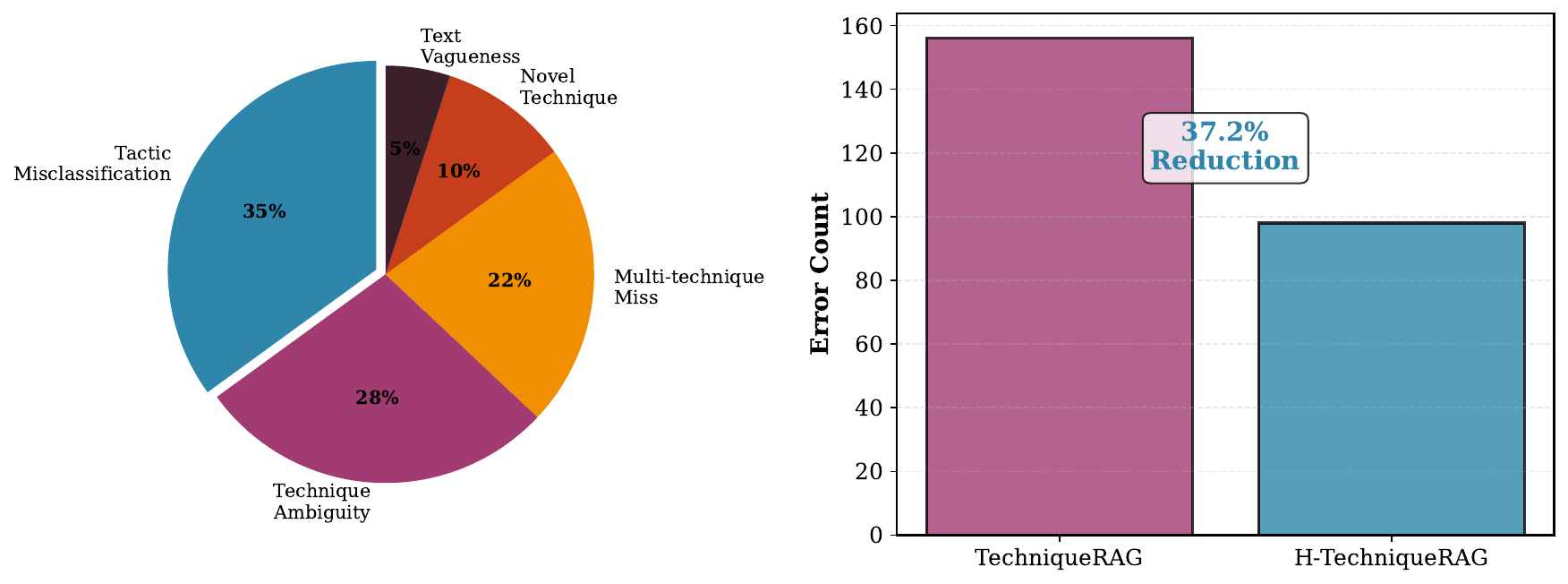}
\vspace{-4mm}
\caption{Error distribution and representative cases.}
\label{fig:error}
\end{figure}

\paragraph{Error Analysis}
We categorize errors into five types (Figure \ref{fig:error}): (1) Tactic Misclassification (35\%): incorrect tactic retrieval leads to wrong technique candidates; (2) Technique Ambiguity (28\%): semantically similar techniques are hard to distinguish; (3) Multi-technique Miss (22\%): complex texts contain multiple techniques, but some are missed; (4) Novel Technique (10\%): techniques not covered in ATT\&CK; (5) Text Vagueness (5\%): insufficient information for any method.

The fallback mechanism mitigates 60\% of tactic misclassification errors by falling back to global retrieval when confidence is low. Technique ambiguity errors could be reduced with more granular technique descriptions or sub-technique modeling.

\section{Conclusion}
\label{sec:conclusion}

We presented H-TechniqueRAG, a hierarchical retrieval-augmented generation framework for CTI-to-ATT\&CK technique annotation. By explicitly leveraging the tactic-technique hierarchy of the ATT\&CK framework, our approach achieves significant improvements in both accuracy and efficiency. The two-stage hierarchical retrieval reduces the candidate set by 77.5\%, while tactic-aware reranking and hierarchical context organization improve annotation precision. Experiments demonstrate consistent improvements across multiple datasets with 62.4\% faster inference and 60\% fewer LLM calls.
The hierarchical approach provides additional benefits beyond performance: interpretable decision paths help analysts understand predictions, and the structured knowledge base enables easier domain adaptation and incremental updates. Our work demonstrates that domain-specific hierarchies, when properly leveraged, can significantly enhance RAG systems.

\bibliography{iclr2025_conference}
\bibliographystyle{iclr2025_conference}

\end{document}